\documentclass[letterpaper, 10 pt, conference]{ieeeconf}

\IEEEoverridecommandlockouts

\usepackage{etextools}
\usepackage{amssymb}
\usepackage{float}
\usepackage{makeidx}
\usepackage{amsmath}
\usepackage{bbm}
\usepackage{epsfig}
\usepackage{epsf}
\usepackage{psfrag}
\usepackage{verbatim}
\usepackage{color}
\usepackage{multirow}
\usepackage{tabularx}
\usepackage{booktabs}
\usepackage[tight,footnotesize]{subfigure}
\usepackage{array}
\usepackage{soul}
\usepackage{footnote}
\usepackage{cite}
\usepackage{dblfloatfix}

\usepackage{color, colortbl}
\usepackage[colorlinks,bookmarksnumbered,citecolor=orange,urlcolor=orange]{hyperref}
\usepackage{graphicx}
\graphicspath{{Figures}}
\DeclareGraphicsExtensions{.pdf,.png}
\usepackage{bigstrut}
\usepackage[english]{babel}

\usepackage{csquotes}

\usepackage[T1]{fontenc}
\usepackage{algorithm, setspace}
\usepackage{algpseudocode}
\usepackage{url}
\usepackage{multirow}
\usepackage{float}
\usepackage{xcolor}
\usepackage{hyperref}
 \hypersetup{
     colorlinks=true,
     linkcolor=orange,
     filecolor=orange,
     citecolor=orange,      
     urlcolor=orange,
     }

\newcommand{\p}[1]{\smallskip \noindent \textbf{{#1}.}}
\newcommand{\eq}[1]{Equation~(\ref{eq:#1})}
\newcommand{\fig}[1]{Figure~\ref{fig:#1}}

\usepackage{balance}

\title{\LARGE

Using High-Level Patterns to Estimate \\ How Humans Predict a Robot will Behave

}

\author{Sagar Parekh$^1$, Lauren Bramblett$^2$, Nicola Bezzo$^2$, and Dylan P. Losey$^1$
\thanks{This work was supported by USDA NIFA Grant
2022-67021-37868 and the Northrop Grumman Corporation's University Basic Research Program.}
\thanks{$^1$Collaborative Robotics Lab (\href{https://collab.me.vt.edu/}{Collab}), Dept. of Mechanical Engineering, Virginia Tech, Blacksburg, VA 24061.
\newline
$^2$Autonomous Mobile Robots Lab (\href{https://www.bezzorobotics.com/}{AMR Lab}), Dept. of Systems \& Information Engineering, Dept. of Electrical \& Computer Engineering, University of Virginia, Charlottesville, VA 22903.
\newline
{Corresponding e-mail: \texttt{sagarp@vt.edu}}}
}

\begin{document}
\maketitle

%%%%%%%%%%%%%%%%%%%%%%%%%%%%%%%%%%%%%%%%%%%%%%%%%%%%%%%%%%%%%%%%%%%%%%%%%%%%%%%%
\begin{abstract}

% When a human interacts with a robot, the human often forms predictions of what the robot will do next.
Humans interacting with robots often form predictions of what the robot will do next.
For instance, based on the recent behavior of an autonomous car, a nearby human driver might predict that the car is going to remain in the same lane.
It is important for the robot to understand the human's prediction for safe and seamless interaction: e.g., if the autonomous car knows the human thinks it is not merging --- but the autonomous car actually intends to merge --- then the car can adjust its behavior to prevent an accident.
Prior works typically assume that humans make precise predictions of robot behavior.
However, recent research on human-human prediction suggests the opposite: humans tend to approximate other agents by predicting their \textit{high-level} behaviors.
We apply this finding to develop a second-order theory of mind approach that enables robots to estimate how humans predict they will behave.
To extract these high-level predictions directly from data, we embed the recent human and robot trajectories into a discrete latent space.
Each element of this latent space captures a different type of behavior (e.g., merging in front of the human, remaining in the same lane) and decodes into a vector field across the state space that is consistent with the underlying behavior type.
We hypothesize that our resulting high-level and course predictions of robot behavior will correspond to actual human predictions.
We provide initial evidence in support of this hypothesis through proof-of-concept simulations, testing our method's predictions against those of real users, and experiments on a real-world interactive driving dataset.

\end{abstract}
%%%%%%%%%%%%%%%%%%%%%%%%%%%%%%%%%%%%%%%%%%%%%%%%%%%%%%%%%%%%%%%%%%%%%%%%%%%%%%%%

\section{Introduction}

Thousands of vehicle crashes caused by driver distractions are reported every year in the United States.
In 2022, for example, distracted drivers were involved in $11,605$ crashes leading to $12,638$ fatalities \cite{nhtsa_dataset}. 
Autonomous and semi-autonomous cars present a potential solution to mitigate human errors and avoid crashes. 
However, these intelligent vehicles need sophisticated decision-making algorithms to safely share the road with pedestrians and human drivers. 

Consider an autonomous car (i.e., a \textit{robot}) driving on a highway alongside a human vehicle.
To understand how the human will behave, existing works focus on inferring the human's objective or intent \cite{dutta2016predicting,hoffman2024inferring}.
For example, is the human an aggressive or a defensive driver? Is the human trying to merge into the robot's lane?
Although estimating the human's intent offers some insight into how the human will interact, it does not provide a complete picture.
During interactions, human reasons about not only their own objective, but also their prediction of how the robot will behave \cite{mutlu2016cognitive}.
Returning to our driving example: if the human thinks the robot is accelerating to pass them, the human may wait and then merge behind them.
Accordingly, --- for the robot to fully understand how the human will behave --- \textit{the robot needs to estimate how the human predicts the robot will behave}.

\begin{figure}
    \centering
    \includegraphics[width=1\columnwidth]{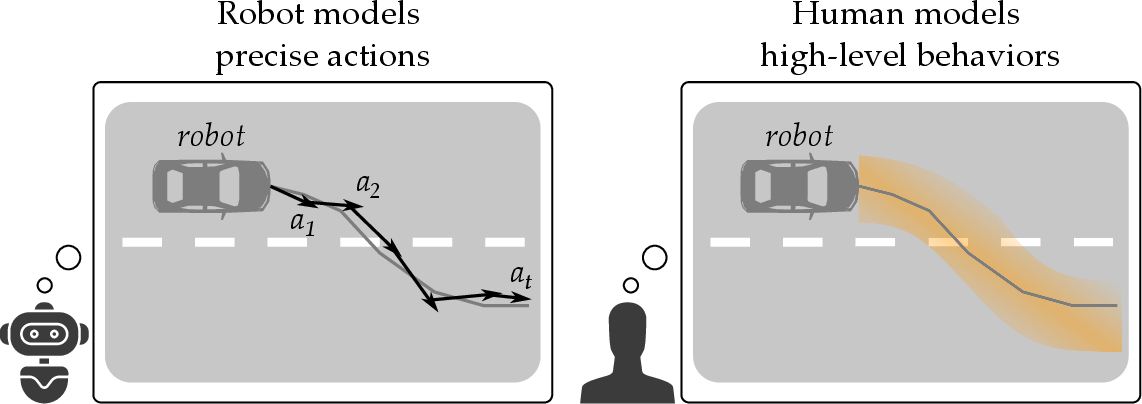}
    \caption{When robots model the behaviors of other robots, they often develop precise predictions. But when humans try to predict the behaviors of robots, they are generally \textit{not} precise. Instead, recent work suggests that humans focus on the high-level pattern in robot behavior (i.e., merging), and then make coarse predictions consistent with that pattern. In this paper we develop a data-driven approach that captures this high-level reasoning, resulting in more accurate estimates of human predictions.}
    \label{fig:front}
    \vspace{-2em}
\end{figure}

This concept of reasoning over another agent's mental model is known as Theory of Mind (ToM) \cite{TOM}.
Specifically, in this paper we will focus on building a second-order ToM of the human (e.g., \textit{"the robot thinks about how human predicts the robot will act"}). 
Existing works often focus only on a first-order ToM: modeling the human's goal so they can provide better assistance \cite{ananth}, estimating the human's strategy in order to influence behavior \cite{lili,rili}, or modeling the human's intent to improve collaboration between humans and robots \cite{ToM2,yuan2021emergence}. Prior research has also explored building second-order ToM for robots; for instance, \cite{sadigh2016planning} models how robot actions affect a human's behavior and leverage this to develop efficient and influential robot policies.
When applied to our example, these works might expect the human to precisely predict each individual action of the autonomous car (see \fig{front}, left).
However, recent research on human-human interaction suggests that humans do not precisely anticipate the actions of other agents \cite{mattson2014superior}; instead, humans appear to reason over more general and intuitive patterns of behavior.
We therefore hypothesize that:
\begin{center}
    \emph{Humans identify high-level behaviors from robots movements, and leverage this high-level inference to approximately predict the robot's next actions.}
\end{center}
\fig{front} highlights this difference.
Existing works often treat humans as if they were robots, and expect the human to predict the individual accelerations and steering angles of the autonomous car.
By contrast, we assume that humans look for repeated patterns of behavior (e.g., going straight, merging lanes), and then classify the robot's actions under one of these high-level patterns (see \fig{front}, right).

In this paper, we apply our hypothesis to create an algorithmic, data-driven method to estimate how humans predict a robot will behave.
Specifically, we develop an autoencoder that inputs recent human and robot behaviors and embeds these trajectories into a discrete latent space (Section~\ref{sec:method}).
This learned latent space (a) inherently forces an information bottleneck that leads to coarse, human-like predictions, and (b) classifies the input trajectories into the discrete, high-level behaviors that were observed in the training dataset.
We show, these discrete latent values often correspond to human-friendly patterns: e.g., merging in front of the human, remaining in the same lane, or merging behind the human (Section~\ref{sec:sim}).
When we decode these discrete latent representations, the robot obtains a vector field of predictions over the environment's state space.
In practice, this enables the robot to estimate how the human thinks it will move right now (in its current state) and also during future timesteps (at other environment states).
Our initial user study suggests that the second-order ToM predictions obtained using this vector field better align with actual human predictions than existing baselines (Section~\ref{sec:user}). 
Additionally, we validate our model's ability to capture high-level movement patterns across a large, real-world dataset for driving domains (Section~\ref{sec:dataset}).

Overall, we make the following contributions:

\p{Formulating Human Predictions of Robot Behavior} 
We formalize our second-order ToM setting, where the human reasons over their own intent and their prediction of the robot's behavior.
The human knows their prediction but not the robot's true policy; conversely, the robot knows its policy, but not how the human predicts it will behave.

\p{Extracting High-Level Predictions} 
We develop a representation learning approach for extracting high-level patterns of behavior.
This approach leverages a discrete latent space to classify the current interaction into intuitive, human-friendly behavior (e.g., driving straight, merging).
We then decode this latent value to obtain a vector field of predictions that the robot can use to estimate how the human thinks it will behave during the rest of the interaction.

\p{Comparing our Model to User Predictions}
We test the quality of our predictions against actual human users.
Across a driving environment and an obstacle course, we find that participants' predictions of robot behavior are more closely aligned with our method than with existing baselines.

\p{Testing on Real-World Agent Interactions}
We evaluate our proposed method using a real-world interactive driving dataset. The experimental results demonstrate that our approach effectively predicts the robot's movement by extracting high-level, human-friendly behavior patterns.
\section{Related Work}

We focus on estimating human predictions of robot behavior.
This problem is rooted in theory of mind, where agents recursively model each other's intent.

\p{Modeling Humans} 
Existing research in human-robot interaction has explored how robots can model human behavior.
These works typically focus on estimating the human's goal, objective, or intent.
For example, in shared autonomy robots infer the human's goals \cite{goals1,goals2}, their intentions \cite{ananth}, even their skills \cite{skill1} in order to assist humans. Alternatively, in multi-agent interaction robots form high-level representations of humans \cite{lili,rili2,pmlr-v205-he23a} to model their behavior (e.g., inferring whether human are aggressive or defensive drivers). 
Overall, these works form \textit{first-order} models of the human (i.e., they estimate what the human is thinking).
But they do not consider how the human reasons over the robot --- i.e., how the human expects the robot to behave.

\p{Theory of Mind} 
Existing literature indicates that humans predict the behavior of other agents during interaction \cite{yoshida2008game,doshi,baker2017rational,bartlet_tom}.
Accordingly, for robots to accurately model humans, we need a \textit{second-order} theory of mind (ToM): the robot must not only infer the human's objective, but also reason over what the human thinks the robot is going to do.
Several works have implemented ToM in robotics \cite{mavrogiannis2019effects,baker2014modeling,brooks2019building}, and robots that use ToM are generally better collaborators than those who do not \cite{ToM2,yuan2021emergence,mrs2}. 
However, current works are often limited to first-order ToM \cite{machineToM,NN-TOM,ToM2,bayesiantom}, or consider second-order ToM only in goal-reaching tasks with additional modeling assumptions \cite{brooks2019building,machineToM}. 
By contrast, in this paper we aim to develop a general purpose, data-driven approach for deriving second-order ToM models from human and robot trajectories.

\p{Representation Learning}
Our approach to achieve this second-order ToM is grounded in representation learning.
Representation learning offers a natural way to impose information bottlenecks and extract underlying patterns from offline datasets.
This is commonly achieved through the application of autoencoder architectures \cite{kingma2013auto,vqvae,fsq} that learn a low-dimensional latent space. 
For instance, with autoencoder frameworks, robots can extract skills from complex behaviors \cite{skill-guided,skill-prior} and develop first-order models of a human collaborator \cite{lili,rili2,NEURIPS2021_a03caec5}. 
We will similarly leverage autoencoders to extract high-level, user-friendly representations of how the human expects the robot to behave.
\section{Problem Statement}\label{sec:problem}

We consider settings where one robot is interacting with one human.
The human and robot could be collaborating, competing, or a combination of both.
We assume that the robot has access to a dataset of offline interactions: this dataset includes the behaviors of the human and robot (e.g., data on how the human vehicle and autonomous car drove around each other in the past).
Based on this dataset --- and the human and robot behavior during the current interaction --- the robot tries to model how the human predicts that robot will behave for the rest of the interaction.
Below we formulate the key aspects of this problem.

\p{Two-Player Game} 
Formally, the human and robot are playing a two-player game.
Let $s_i \in \mathcal{S}$ be the state of agent $i$, and let $a_i \in \mathcal{A}$ be the action of agent $i$.
In what follows, we will use $i = 1$ to refer to the robot and $i = 2$ to refer to the human.
Both agents (i.e., the human and robot) have policies that map the system state to actions.
The robot's policy $\pi_1(s) \rightarrow a_1$ determines the robot's actions, and similarly the human's policy $\pi_2(s) \rightarrow a_2$ determines the human's actions.
Within our problem setting the robot knows its own policy, but it \textit{does not know} how the human is modeling the robot's policy.
On the other hand, the human \textit{is unaware} of the robot's policy $\pi_1$, though they have an understanding of their own predictions regarding the robot's behavior.

\p{Trajectories} 
Let $s^t = (s_1^t, s_2^t)$ be the joint state of the human-robot system at timestep $t$, and let $a^t = (a_1^t, a_2^t)$ be their joint action. 
In our driving example, state $s_i$ is the position of agent $i$ in the environment, action $a_i$ is its velocity, and $s$ captures the positions of both agents along the highway.
A trajectory $\xi \in \Xi$ is a sequence of $T$ joint state-action pairs: $\xi = \{(s^0, a^0), \dots, (s^T, a^T)\}$.  

\p{Dataset} 
To develop our model of the human's prediction we assume access to an offline dataset of $N$ trajectories: $\mathcal{D} = \{\xi^1, \xi^2, \dots, \xi^N\}$.
These trajectories could be collected from actual human-robot interactions, or by training two simulated agents and rolling out their interactions.
Intuitively, trajectories may contain patterns of behavior that human's recognize: for example, the trajectory of the autonomous car and human car could correspond to passing, merging, or yielding behaviors.
We will refer to these high-level patterns of behavior as $z \in \mathcal{Z}$.
We emphasize that the robot is not given $z$ \textit{a priori}; instead, the robot must extract these high-level behaviors from the trajectory dataset $\mathcal{D}$.

\section{Extracting High-Level Behaviors for Intuitive Human Predictions} \label{sec:method}

Our objective is to estimate how humans predict a robot will behave. 
In order to characterize these predictions, we utilize our hypothesis: humans, unlike robots, do not make precise, step-by-step predictions of robot motion \cite{mattson2014superior}. 
Instead, humans reason over the robot's motion using high-level patterns (e.g., passing, merging), and then apply these patterns to make coarse estimates of the robot's action.
In this section, we develop a representation learning approach that seeks to emulate this human process.
Specifically, we use an offline dataset of interactions to train an autoencoder that learns a discrete latent space. 
The discrete latent space is designed to introduce an information bottleneck that extracts distinct, high-level patterns of robot behavior. 
Returning to our driving example: one discrete value might correspond to the robot staying in the same lane, while another might capture a robot merging into the human's lane.
In what follows, we provide the details of this approach.
We start with an introduction of the network architecture. 
Next, we describe the procedure for extracting high-level behaviors from this model. 
Finally, we leverage the trained model to estimate a human's predictions of robot actions.

\begin{figure}
    \centering
    \includegraphics[width=1\columnwidth]{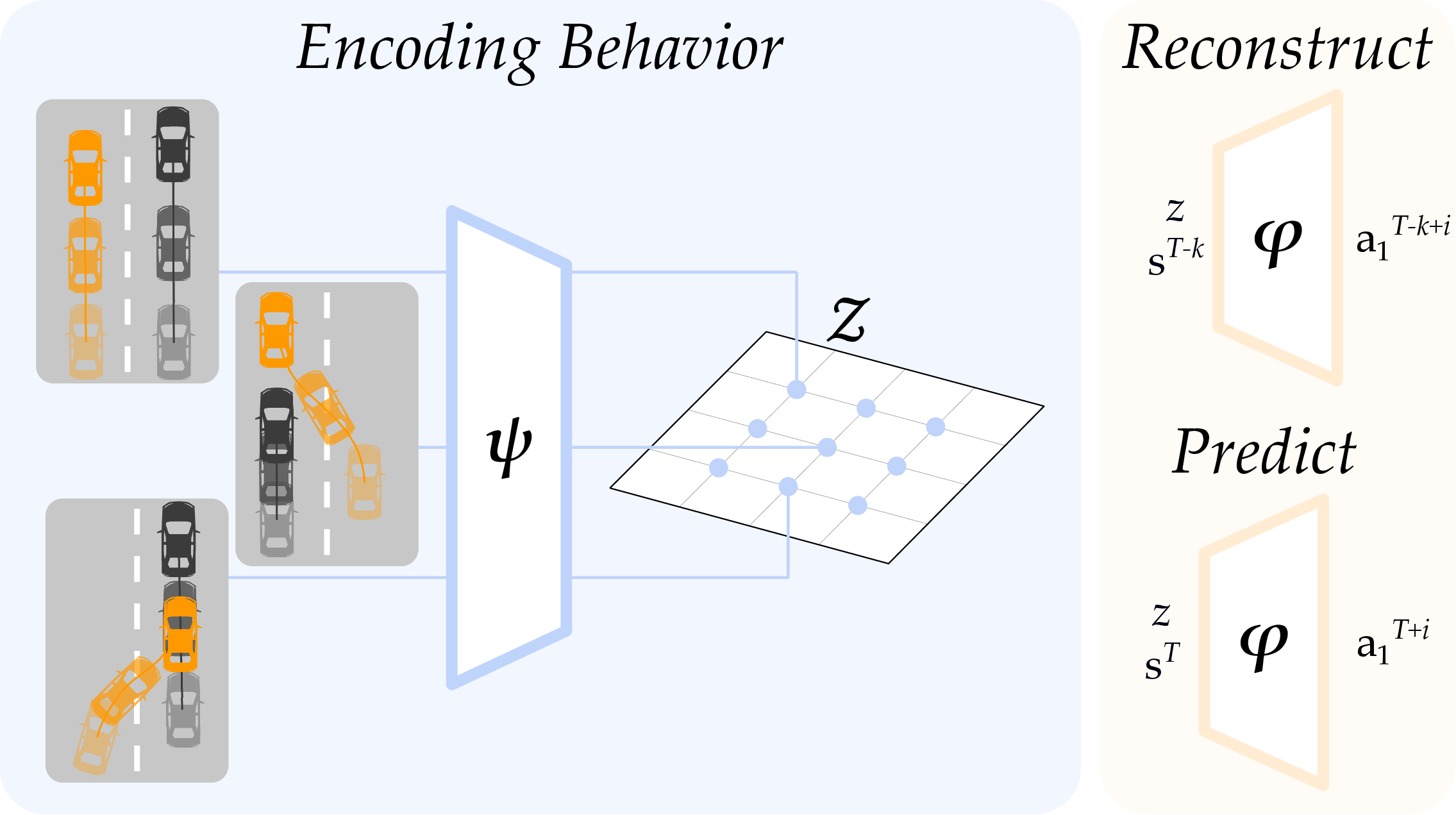}
    \caption{Our proposed model for learning to estimate how humans predict a robot will behave. We develop an autoencoder with three parts: (a) an encoder $\phi$ that inputs trajectories and outputs discrete latent values, (b) a discrete latent space $z$ where each different value corresponds to a different high-level behavior, and (c) a decoder $\psi$ that takes the current state and high-level pattern and predicts the robot's next $n$ actions. During training, the decoder additionally reconstructs action sequences $a_1^{T-k+i}$ for a state $s^{T-k}$ from the input trajectory in order to learn the discrete latent space.}
    \label{fig:method}
    \vspace{-1.5em}
\end{figure}

\p{Model Architecture}
As described in Section \ref{sec:problem}, each trajectory $\xi$ is an interaction between two agents. 
Agents may demonstrate different behaviors during different interactions. 
Accordingly, we want to learn a low-dimensional space of behaviors $\mathcal{Z}$ that can distinctly represent the different types of behaviors demonstrated by the robot. 
To this end, we use a special class of autoencoders with finite scalar quantization \cite{fsq}: instead of a continuous latent space, finite scalar quantization embeds trajectories into a discrete latent space with a fixed number of points.
When trained, each of the discrete latent vectors corresponds to a different class of behavior demonstrated by the robot (see \fig{method}). 
Consider a latent space of dimension $d$, where the autoencoder employs a set of $L$ distinct values within the interval $[-1, 1]$ to quantize each of the $d$ channels.
Each channel of the latent vector can have only one of the $L$ values, and the entire latent space has $L^d$ discrete vectors. 
Mathematically, for a latent vector $z$, each channel $z_i$ is linearly scaled to an integer in the range $[0, L - 1]$ with the equation \cite{fsq}: 
$$z_i^{int} = \text{round}\left(\frac{L - 1}{2}\left(\tanh\left(z_i\right)\right) + 1\right)$$
This integer is then re-scaled to a discrete value within the range $[-1, 1]$ by applying the formula:
$$z_i^q = \frac{2}{L - 1} z_i^{int}$$
In practice, the scaling and re-scaling maps each channel of the embedding $z$ to a single value in $\{-1, -1 + \frac{2}{L - 1}, \dots, 1\}$. 
For example, if we let $d=3$ and $L=2$, then the discrete latent space will contain $2^3 = 8$ vectors. 
Each channel in this vector can have a value of $-1$ or $1$; therefore, the discrete latent space will be a set $\mathcal{Z} = \{(-1, -1, -1), (-1, -1, 1), (-1, 1, 1), \dots, (1, 1, 1)\}$. 

Now that we have described the discrete latent space, we are ready to analyze our overall autoencoder.
This autoencoder consists of an \textit{encoder} network, which inputs a trajectory $\xi$ and embeds it to the discrete latent space $z$, and a \textit{decoder} network, which takes the latent value $z$ and attempts to reconstruct parts of the input trajectory $\xi$.
Ideally, we would train the encoder and decoder networks to accurately reconstruct the trajectory data.
However, using a discrete latent space presents fundamental challenges.
Specifically, since the quantization operation is not differentiable, normal gradient-based optimization techniques --- such as standard backpropagation --- cannot be directly applied.
To solve this problem, we utilize a straight-through estimator technique that copies the gradients computed from the decoder directly to the encoder, allowing the encoder to be updated and refined during optimization \cite{vqvae}.

This solution enables us to take advantage of the quantized latent space.
Importantly, discretization provides a significant bottleneck in the information flow from encoder to decoder (see \fig{method}) by limiting the number of latent vectors available for encoding. 
Consequently, to minimize the loss function, the learned latent space must spread information across the quantized bins: i.e., each latent value should capture a different behavior pattern.
Returning to our driving example, in \fig{method}, three distinct behaviors of the robot  --- driving straight, merging left, merging right --- are mapped to different latent vectors $z \in \mathcal{Z}$.
More generally, quantization of the latent space to discrete values has been shown to avoid posterior collapse --- where the latent values fail to capture meaningful representations \cite{vqvae,fsq,sqvae}.

\p{Encoding High-Level Behaviors} 
So far we have described our overarching autoencoder architecture with a discrete latent space. 
Next, we focus on the details for the encoder and decoder.
The encoder $\psi$ is a fully connected network that maps trajectories to discrete values: $\phi(\xi) \rightarrow z$.
More specifically, this encoder maps the trajectory of the two interacting agents $\xi = \{(s^0, a^0), \dots, (s^T, a^T)\}$ to a quantized latent vector $z \in \mathcal{Z}$ from the discrete latent space.

\p{Decoding High-Level Behaviors to Predict Actions} 
Once we embed the interaction trajectory to a discrete value (i.e., to a high-level behavior), our final step is to convert this high-level behavior into a specific prediction about the robot's actions.
We achieve prediction using the decoder $\phi(z, s^T)$. 
This decoder inputs the latent vector $z$ and $s^T$, the most recent state of the human and robot.
The decoder then maps this current state and the high-level pattern to a sequence of robot actions.
To effectively train our autoencoder model, we actually leverage this decoder \textit{twice}.

First, we use it to \textit{predict} the next $n$ robot actions: $[a_1^{T+1}, a_1^{T+2}, \dots a_1^{T+n}]$.
The error between our model's prediction and the real actions of the robot is:
\begin{align}
    \mathcal{L}_{pred} &= \frac{1}{n}\sum_{k=T+1}^{T+n} \left( a_1^k - \hat{a}_1^k \right)^2 \label{eq:predictionL}
\end{align}
where $\hat{a}_1^k$ is the action at timestep $k$ output by our decoder, and $a_1^k$ is the robot's true action at the same timestep.

Second, we leverage our decoder to \textit{reconstruct} a series of $n$ actions from the input trajectory.
Here we sample a random state $s^t$ from trajectory $\xi$, and then attempt to reconstruct the next $n$ robot actions: $[a_1^{t+1}, \dots, a_1^{t+n}]$.
The error between the true actions and the prediction is:
\begin{align}
    \mathcal{L}_{recon} &= \frac{1}{n}\sum_{k=t}^{t+n} \left( a_1^k - \hat{a}_1^k \right)^2 \label{eq:reconstructionL}
\end{align}
The purpose of this reconstruction loss is to train the latent space.
If each value of $z \in \mathcal{Z}$ corresponds to a different high-level behavior, then passing $z$ and $s^t$ into the decoder should be sufficient to accurately anticipate the rest of the robot's behavior.
For instance, if the encoder embeds a portion of the trajectory to a $z$ that captures an autonomous car merging, then the network can predict how that car will continue to merge for the remainder of the trajectory.

\p{Training} We train our encoder, latent space, and decoder using both losses: the prediction loss \eq{predictionL} and the reconstruction loss \eq{reconstructionL}.
This results in the model shown in \fig{method}, where the encoder takes in a trajectory of recent behavior and maps that trajectory to a high-level pattern.
The decoder then converts the current state and the high-level pattern into a predicted sequence of robot actions.
As a reminder, we recognize that the robot knows its own policy --- the robot could precisely say what actions it will take next.
But the purpose of this model is not to predict the robot's actions accurately; instead, we are trying to develop a prediction that matches how the human reasons over the robot.
We will put our model to the test in the following sections.
First, we will explore the types of high-level behaviors that our network captures, and then second, we will compare the predictions from our network to the predictions of actual users.
\section{Testing the Encoded High-Level Behaviors}\label{sec:sim}

\begin{figure*}
    \centering
    \includegraphics[width=1.7\columnwidth]{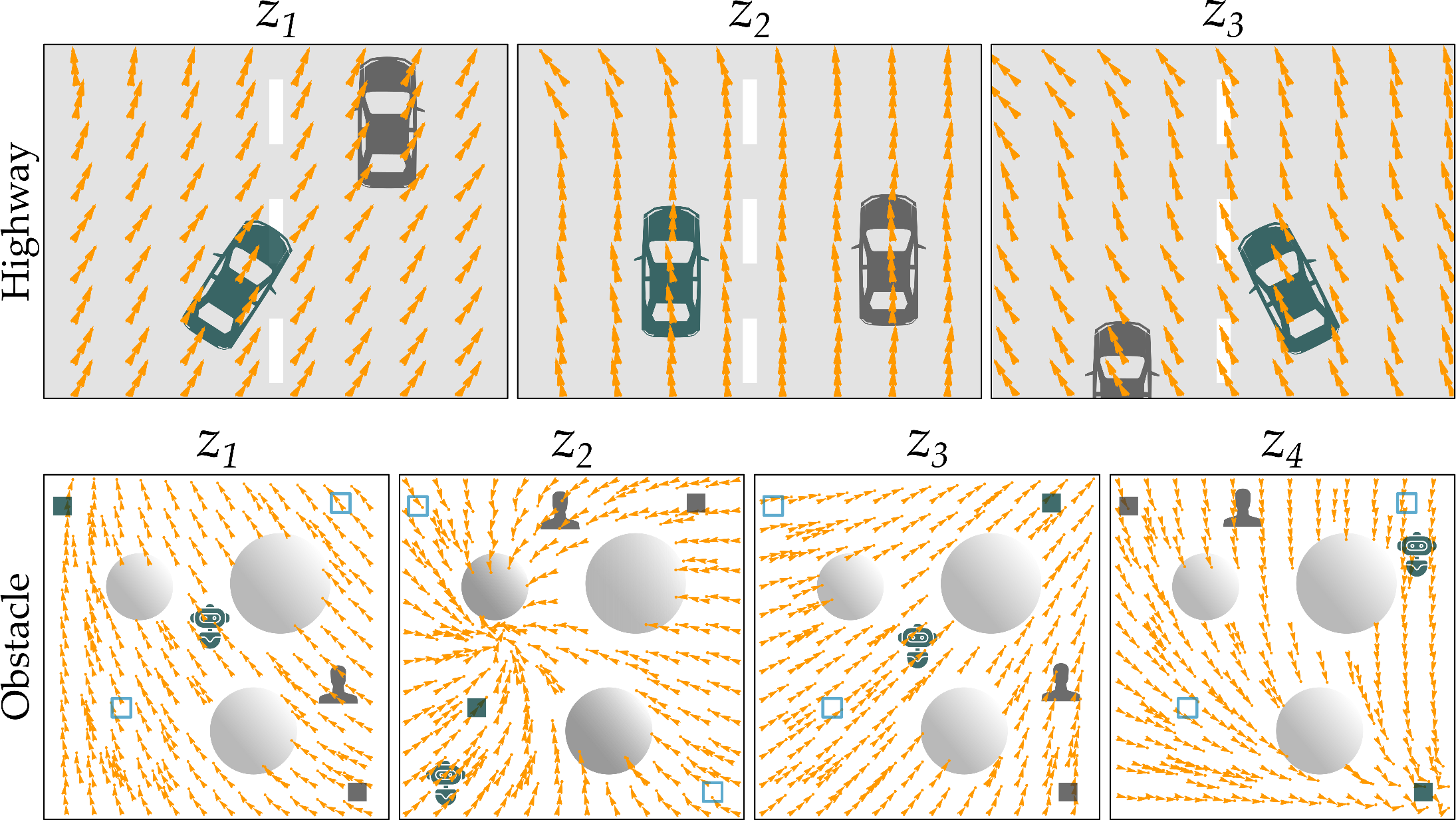}
    \caption{Human predictions of robot behavior extracted by our method. The human is shown in gray, and the robot is shown in green. Our approach first embeds the current interaction into a discrete representation $z \in \mathcal{Z}$. We then fix that value of $z$, and extract the actions we think the human will predict across the workspace. This results in a vector field that visualizes the high-level pattern $z$. (Top) In the \textit{Highway} environment, our method autonomously extracts three high-level behavior patterns of the robot: merging into the right lane, staying straight, and merging into the left lane. (Bottom) In \textit{Obstacle}, our method identifies the goal-reaching movement patterns of the robot, where each different $z$ predicts actions that move towards a specific goal. These results suggest that our data-driven approach is able to learn high-level behaviors that align with human explanations (e.g., merging, going to the left goal).}
    \label{fig:sim_results}
    \vspace{-0.5em}
\end{figure*}

In this section we explore the types of high-level behaviors that are extracted by our approach.
Here we do not use actual human data; instead, we start with synthetic interactions, and study whether the underlying patterns our algorithm recovers from these interactions match with intuitive human interpretations.
We consider two multi-agent environments with a human and robot: \textit{Highway} and \textit{Obstacle}. 

\p{Environments} 
In the \textit{Highway} environment one human-driven vehicle is on the same multi-lane road as an autonomous car (see \fig{sim_results}, top row). 
Both cars could remain in their current lanes, the human could merge into the robot's lane, or the robot could merge into the human's lane.
In the \textit{Obstacle} environment the two agents move towards their respective goals while avoiding static obstacles (see \fig{sim_results}, bottom row). 
There are four total goals in the environment, and the goals assigned to the human and robot are chosen randomly at the start each each new interaction.

\p{Training} 
For both environments we collect a dataset of trajectories $\mathcal{D}$. 
In order to explore the types of patterns learned by our system, in this section we generate these trajectories using \textit{simulated} human and robot data.
Specifically, we give the simulated human and robot reward functions, and then sample trajectories that optimize these reward functions using model predictive control (in \textit{Highway}) or the soft-actor critic algorithm (in \textit{Obstacle}) \cite{sac}.
Overall, in the \textit{Highway} environment we collect $2000$ synthetic trajectories, and in the \textit{Obstacle} environment we generate $2000$ trajectories. 
Given this dataset $\mathcal{D}$ of human and robot interactions, we train our method using the method described in Section~\ref{sec:method}.

\p{Visualizing High-Level Behaviors} 
Our method embeds the human-robot trajectories into a discrete latent space.
Based on the properties of our model structure, this latent space should theoretically cluster different interaction patterns within different latent vectors.
In \fig{sim_results} we show experimental evidence to support this distinct clustering.

To obtain this figure we first trained our method on the offline dataset, and then sampled a grid of states throughout the environment.
At every state we visualized the action (i.e., human prediction) output by our decoder using each discrete value of $z$.
For example, in the \textit{Highway} environment we selected latent values $z_1$, $z_2$, and $z_3$, and then decoded each state with those latent values to obtain three different vector fields.
Examining the results, we notice that these vector fields align with intuitive explanations of robot behavior.
Within the \textit{Highway} environment (top row of \fig{sim_results}) the vector fields appear to correspond to the robot merging into the right lane, the robot going straight, and the robot merging into the bottom lane.
Similarly, in the \textit{Obstacle} environment (bottom row of \fig{sim_results}) the vector fields approximately capture how the robot might move to different goal locations.
Overall, these simulation results suggest that our autoencoder design is able to extract distinct high-level patterns, and that these high-level patterns align with how humans might interpret the robot's behavior.

\section{Comparing to User Predictions} \label{sec:user}

Our ultimate goal is to enable a robot to estimate how humans predict it will behave. 
In Section \ref{sec:sim} we qualitatively found that our proposed approach was able to identify high-level behaviors demonstrated by the robot. 
However, so far we have only worked with simulated data.
Now in Section~\ref{sec:user} we put our predictive algorithm to the test with \textit{actual humans}.
We validate whether the robot's extracted behaviors align with those that real participants attribute to the robot.

\begin{figure*}
    \centering
    \includegraphics[width=1.8\columnwidth]{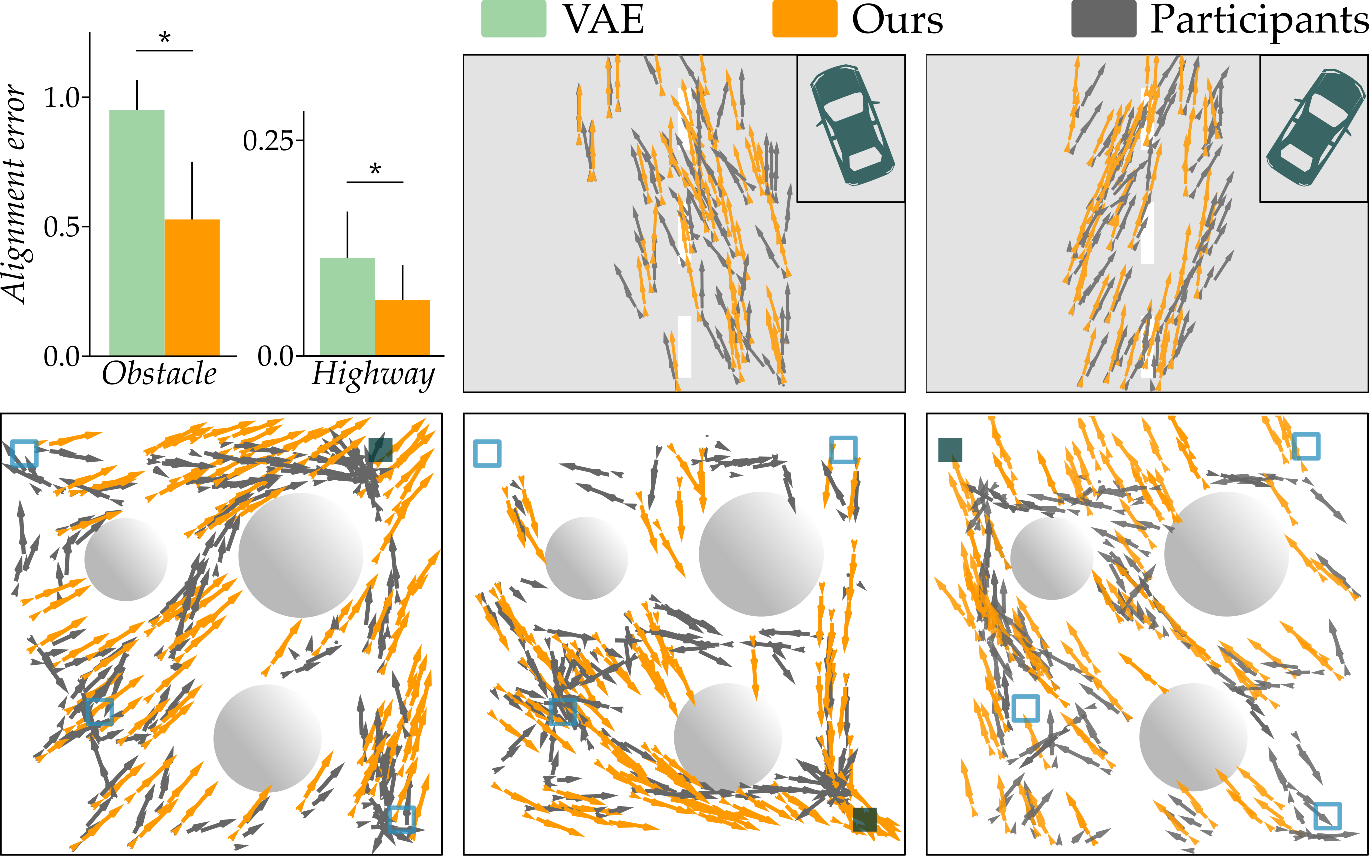}
    \caption{Results from our user study. Participants observe a segment of human-robot motion and predict how the robot will behave. We compare these real predictions to the predictions made by our method and the baseline. 
    % The plot shows the mean \textit{alignment error} of both methods. The error value can range from $0$ (indicating parallel prediction) to $2$ (indicating opposite predictions). Our method achieves a significantly lower error than the baseline in both environments, \textit{Highway} and \textit{Obstacle}
    (Top-left) The plot shows the mean \textit{alignment error} of both methods. The error value can range from $0$ (indicating parallel prediction) to $2$ (indicating opposite predictions). Our method achieves a significantly lower error than the baseline in both environments, \textit{Highway} and \textit{Obstacle}. Next, we compare the vector field of the participants' predictions with the vector field decoded from the most common latent vectors. (Top-right) In \textit{Highway}, we obtain two prominent movement patterns: the robot merging left, and the robot merging right. The predictions made by our model (orange) align with the participants' predictions. (Bottom) In \textit{Obstacle}, we obtain three notable goal-reaching behaviors. The behaviors decoded from each latent clearly indicates the robot's goal, and this appears to align with the participant's model of the robot.}
    \label{fig:study_results}
    \vspace{-0.5em}
\end{figure*}

\p{Independent Variables} We compare our method (\textbf{Ours}) with a standard variational autoencoder (\textbf{VAE}) \cite{kingma2013auto}. 
\textbf{VAE} is a commonly used architecture to extract high-level information about an agent from its trajectories \cite{skill-guided,rili2,NEURIPS2021_a03caec5}.
Both of these methods map the robot's trajectories into a low-dimensional representation which extracts the robot's high-level behaviors.
However, the structure of this representation space is different for the two architectures ---
while \textbf{VAE} learns a Gaussian posterior over a continuous space, \textbf{Ours} learns a finite, discrete space.
The discretization allows our method to extract intuitive, user-friendly behaviors from interactions, while the VAE learns more precise, robot-like estimates.
Both the methods are pre-trained on the same dataset of interactions $\mathcal{D}$ as explained in Section \ref{sec:sim}.

\p{Dependent Measures} We evaluate our method using objective and subjective measures. 
First, we assess how closely the actions predicted by each model align with those made by participants by calculating the cosine distance, referred to as the \textit{alignment error}. 
The alignment error ranges from $0$ to $2$, where $0$ indicates perfect alignment between model predictions and participant predictions, and $2$ indicates completely opposite predictions.

Next, we validate that the high-level behaviors extracted by our method correspond to the high-level behaviors attributed to the robot by humans. 
To this end, we compare the predictions associated with different latent vectors in our discrete representation space to those made by the participants, ensuring that the learned behaviors align with human expectations.

\p{Experiment Setup} Participants were asked to predict the movement of a robot in the two environments introduced in Section \ref{sec:sim}: \textit{Highway} and \textit{Obstacle}.
In the \textit{Highway} environment, two agents --- the robot and human --- drive in parallel lanes alongside each other. In the \textit{Obstacle} environment, the two agents navigate toward their respective goals.
Participants were shown a brief segment of the two agents' movement over a span of $5-7$ timesteps.
Based on this observation, participants were then asked to predict the robot’s future actions over the next $3$ timesteps.

\p{Participants and Procedure} We conducted an online user study with $22$ participants recruited from the Virginia Tech and University of Virginia communities. 
Every participant completed $10$ trials in each of the environments. 
In each trial, they observed 4 full interactions, followed by a partial segment of an interaction.
After viewing these trajectories, participants predicted the robot’s next $3$ actions, starting from $5$ different initial states, denoted $s_1$.
Additionally, both \textbf{Ours} and \textbf{VAE} were used to map the interaction segment to a high-level behavior $z$ within the model's representation space.
The decoder in each model then predicted the actions for the $5$ states using the corresponding $z$.

\p{Hypothesis} Our hypothesis \textbf{H} is that:
\begin{displayquote}
    \textit{\textbf{Ours} will more accurately match the predictions made by human participants as compared to \textbf{VAE}.}
\end{displayquote}

\p{Results} Our results are summarized in \fig{study_results}. 
The plots in the top-left illustrate the average \textit{alignment error} across all participants in both environments. 
We observe that \textbf{Ours} achieves a lower \textit{alignment error} compared to \textit{VAE} in  both the \textit{Highway} and \textit{Obstacle}.
A paired t-test confirms that these differences in the error are statistically significant ($p < 0.01$).
This demonstrates that our method is more accurate at estimating how participants predicts the robot will move, and supports our hypothesis.

\begin{figure*}
    \centering
    \includegraphics[width=1.8\columnwidth]{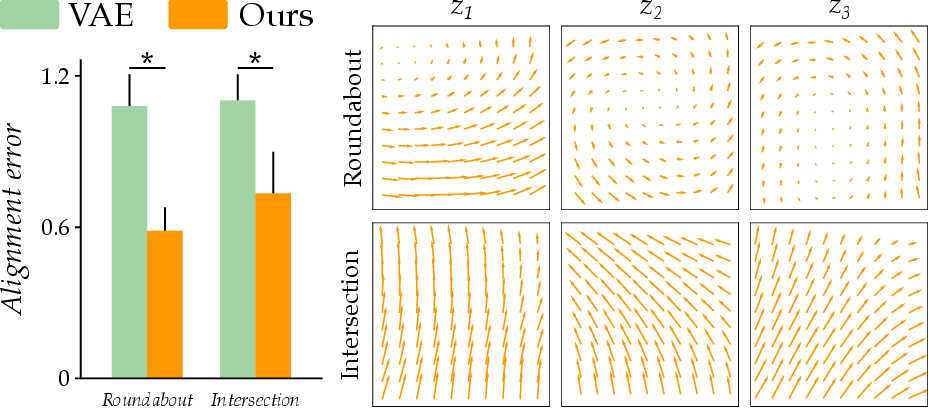}
    \caption{Results of our experiments on the INTERACT dataset \cite{interactiondataset}. We observe the trajectory of two cars, and predict the movement of one vehicle from the perspective of the other car. We compare the predictions of our high-level method and the precise baseline. (Left) The average \textit{alignment error} of the predictions across $10$ trials. The left plot shows the results for the \textit{Roundabout} scenario and the right shows the results for the \textit{Intersection} scenario. The error value ranges from 0 (indicating parallel prediction) to 2 (indicating opposite prediction). Our method achieves a significantly lower error in both scenarios ($p < 0.001$). Next, we visualize the movement patterns identified by the latent vectors of our model. (Top-right) In \textit{Roundabout} we obtain three prominent movement patterns: entering the roundabout from the left and going up, going around the roundabout, and entering the roundabout from the bottom. (Bottom-right) In \textit{Intersection} our model identifies three prominent movement patterns: going straight, going left, and going right.}
    \label{fig:dataset_results}
    \vspace{-0.5em}
\end{figure*}
\section{Testing on a Real-World Dataset} \label{sec:dataset}

In Section \ref{sec:user}, we validated our approach with data from $22$ participants in custom environments.
But just because our approach was able to accurately predict how humans thought a robot would behave in these custom environments does not mean that our approach will work across general, real-world interactions.
To more broadly test our approach, in Section~\ref{sec:dataset} we leverage an existing dataset that includes driving motions from users around the world.
We again explore whether our high-level prediction method is able to infer how humans think other agents will behave.

\p{Dataset}
We used the INTERACTION dataset \cite{interactiondataset} for our analysis. 
This dataset consists of traffic across a variety of interactive real-world scenarios. 
Specifically, the dataset provides $4$ second trajectories of cars driving in different road settings collected from $11$ locations.
We studied two interactive driving scenarios from this dataset: \textit{Roundabout} and \textit{Intersection}.
Each trajectory is composed of a sequence of $(x, y)$ positions and velocities for the cars.

% \begin{figure}[h]
%     \centering
%     \includegraphics[width=0.5\columnwidth]{Figures/dataset_results.eps}
%     \caption{Results of our experiments on the INTERACT dataset. We observe the trajectory of two cars and model the movement of an ego agent and predict its future action. We compare the predictions of our method and the baseline. The plot shows the mean \textit{alignment error} of the predictions across $10$ trials. The left plot shows the results for the \textit{Roundabout} scenario and the right shows the results for the \textit{Intersection} scenario. The error value ranges from 0 (indicating parallel prediction) to 2 (indicating opposite prediction). Our method achieves a significantly lower error in both scenarios ($p < 0.001$).}
%     \label{fig:dataset_results}
% \end{figure}

\p{Procedure}
The INTERACTION dataset provides training and validation sets.
We used the training set to learn a model of the agent's behaviors using our method (\textbf{Ours}) and the baseline (\textbf{VAE}).
In each trajectory, we designate one agent as the ``robot,'' (i.e., the agent whose movement we want to predict), and we designate the other agent as the human.
The models observe a $1$ second segment of the trajectory and predict the robot's next action conditioned on the two agents' states.
We applied the \textit{alignment error} defined in Section~\ref{sec:user} to compare the performance of the two methods.

\p{Results}
The plot in \fig{dataset_results} (left) shows the mean \textit{alignment error} across $10$ trials. The \textit{alignment error} for \textbf{Ours} is consistently lower than that of the baseline \textbf{VAE} in both scenarios, \textit{Roundabout} and \textit{Interaction}. A paired t-test confirms that these differences in the error are statistically significant ($p < 0.001$), indicating that our method more accurately models the robot's behavior and predicts it's movements. This finding supports our hypothesis \textbf{H}. 

To better understand why our approach was able to generate predictions that better aligned with the actual human's prediction, we analyzed the high-level patterns extracted by our method.
See \fig{dataset_results} (right).
As a reminder, our goal is to enable the robot to understand how a human thinks it will behave.
With the \textbf{VAE} baseline, the robot treats the human as if it were a robot; i.e., the robot thinks the human will make precise, highly accurate predictions.
But with \textbf{Our} approach the robot recognizes that human's reason over higher-level models --- instead of precise predictions, the human thinks about how the autonomous car might change lanes or exit the roundabout.
Our method autonomously extracts these higher-level predictions in \fig{dataset_results}.
For instance, in the \textit{Roundabout} scenario the latent vector $z_1$ represents the movement of entering the roundabout from the left and heading upwards, while the latent vector $z_2$ captures the behavior of navigating around the entire roundabout. 
Similarly, in the \textit{Intersection} scenario, different latent vectors correspond to distinct movement patterns: going straight ($z_1$), going left ($z_2$), or going right ($z_3$).
These high-level patterns explain why our method is able to more accurately reflect the human predictions across this real-world dataset.

\p{Limitations and Future Work} While our method yields promising results, it does not perfectly cluster distinct interaction patterns into separate latent vectors.
This is evident from the second latent vector $z$ in the \textit{Obstacle} environment (\fig{study_results}): our model predicts the robot moving toward the bottom-right goal, yet participants' predictions sometimes indicate movement towards the bottom-left goal.
We hypothesize that this limitation could be addressed by incorporating priors or added supervision that captures the types of high-level behaviors humans expect the system to extract.

Moving forward, our future work will incorporate the human's prediction of the robot's motion into the robot's decision making.
With our method, the robot now has a data-driven way to express how nearby humans predict that robot will behave.
A ToM robot should then reason over this prediction when making its decisions; for instance, if an autonomous car knows that nearby humans expect it to exit the roundabout, and it actually wants to stay on the roundabout, then it should take additional actions to make its intention clear to the humans.
\section{Conclusion}

We have presented a theory of mind approach that enables a robot to estimate how humans predict it will behave.
Unlike previous research --- which often assumes that humans will precisely infer the robot's policy --- we recognized that humans typically rely on course, high-level predictions.
For instance: instead of predicting the exact speed and velocity of an autonomous car, the human might reason over whether that car is passing or merging.
To extract these high-level predictions directly from interaction data we introduced an autoencoder with a discrete latent space. 
This latent space served as an information bottleneck (providing course predictions), and its discrete structure yielded different types of behavior (capturing high-level patterns).
We tested the feasibility of our method to accurately estimate human predictions through i) proof-of-concept simulations, ii) a user study with real human predictions, and iii) tests across a large, real-world interactive driving dataset.
Our results suggest that separating the robot's behavior into high-level clusters leads to more accurate estimates than precise baselines.

%%%%%%%%%%%%%%%%%%%%%%%%%%%%%%%%%%%%%%%%%%%%%%%%%%%%%%%%%%%%%%%%%%%%%%%%%%%%%%%%%

\balance
\bibliographystyle{IEEEtran}
\bibliography{IEEEabrv,citations}

\begin{thebibliography}{10}
\providecommand{\url}[1]{#1}
\csname url@rmstyle\endcsname
\providecommand{\newblock}{\relax}
\providecommand{\bibinfo}[2]{#2}
\providecommand\BIBentrySTDinterwordspacing{\spaceskip=0pt\relax}
\providecommand\BIBentryALTinterwordstretchfactor{4}
\providecommand\BIBentryALTinterwordspacing{\spaceskip=\fontdimen2\font plus
\BIBentryALTinterwordstretchfactor\fontdimen3\font minus \fontdimen4\font\relax}
\providecommand\BIBforeignlanguage[2]{{%
\expandafter\ifx\csname l@#1\endcsname\relax
\typeout{** WARNING: IEEEtran.bst: No hyphenation pattern has been}%
\typeout{** loaded for the language `#1'. Using the pattern for}%
\typeout{** the default language instead.}%
\else
\language=\csname l@#1\endcsname
\fi
#2}}

\bibitem{nhtsa_dataset}
\BIBentryALTinterwordspacing
{National Highway Traffic Safety Administration}, ``Crash report sampling system (crss),'' 2022. [Online]. Available: \url{https://cdan.dot.gov/databook/databook.htm#}
\BIBentrySTDinterwordspacing

\bibitem{dutta2016predicting}
V.~Dutta and T.~Zielinska, ``Predicting the intention of human activities for real-time human-robot interaction (hri),'' in \emph{Social Robotics}, 2016, pp. 723--734.

\bibitem{hoffman2024inferring}
G.~Hoffman, T.~Bhattacharjee, and S.~Nikolaidis, ``Inferring human intent and predicting human action in human--robot collaboration,'' \emph{Annual Review of Control, Robotics, and Autonomous Systems}, 2024.

\bibitem{mutlu2016cognitive}
B.~Mutlu, N.~Roy, and S.~{\v{S}}abanovi{\'c}, ``Cognitive human--robot interaction,'' \emph{Springer Handbook of Robotics}, pp. 1907--1934, 2016.

\bibitem{TOM}
S.~Baron-Cohen, A.~M. Leslie, and U.~Frith, ``Does the autistic child have a “theory of mind”?'' \emph{Cognition}, pp. 37--46, 1985.

\bibitem{ananth}
A.~Jonnavittula and D.~P. Losey, ``I know what you meant: Learning human objectives by (under)estimating their choice set,'' in \emph{International Conference on Robotics and Automation}, 2021, p. 2747–2753.

\bibitem{lili}
A.~Xie, D.~Losey, R.~Tolsma, C.~Finn, and D.~Sadigh, ``Learning latent representations to influence multi-agent interaction,'' in \emph{Conference on Robot Learning}, 2021, pp. 575--588.

\bibitem{rili}
S.~Parekh, S.~Habibian, and D.~P. Losey, ``Rili: Robustly influencing latent intent,'' in \emph{International Conference on Intelligent Robots and Systems}, 2022, pp. 01--08.

\bibitem{ToM2}
T.~X. Lim, S.~Tio, and D.~C. Ong, ``Improving multi-agent cooperation using theory of mind,'' \emph{arXiv preprint arXiv:2007.15703}, 2020.

\bibitem{yuan2021emergence}
L.~Yuan, Z.~Fu, L.~Zhou, K.~Yang, and S.-C. Zhu, ``Emergence of theory of mind collaboration in multiagent systems,'' \emph{arXiv preprint arXiv:2110.00121}, 2021.

\bibitem{sadigh2016planning}
D.~Sadigh, S.~Sastry, S.~A. Seshia, and A.~D. Dragan, ``Planning for autonomous cars that leverage effects on human actions.'' in \emph{Robotics: Science and Systems}, 2016, pp. 1--9.

\bibitem{mattson2014superior}
M.~P. Mattson, ``Superior pattern processing is the essence of the evolved human brain,'' \emph{Frontiers in neuroscience}, p. 265, 2014.

\bibitem{goals1}
S.~Javdani, H.~Admoni, S.~Pellegrinelli, S.~S. Srinivasa, and J.~A. Bagnell, ``Shared autonomy via hindsight optimization for teleoperation and teaming,'' \emph{The International Journal of Robotics Research}, pp. 717--742, 2018.

\bibitem{goals2}
H.~J. Jeon, D.~P. Losey, and D.~Sadigh, ``Shared autonomy with learned latent actions,'' \emph{arXiv preprint arXiv:2005.03210}, 2020.

\bibitem{skill1}
N.~Enayati, G.~Ferrigno, and E.~De~Momi, ``Skill-based human--robot cooperation in tele-operated path tracking,'' \emph{Autonomous Robots}, pp. 997--1009, 2018.

\bibitem{rili2}
S.~Parekh and D.~P. Losey, ``Learning latent representations to co-adapt to humans,'' \emph{Autonomous Robots}, pp. 771--796, 2023.

\bibitem{pmlr-v205-he23a}
J.~Z.-Y. He, Z.~Erickson, D.~S. Brown, A.~Raghunathan, and A.~Dragan, ``Learning representations that enable generalization in assistive tasks,'' in \emph{Conference on Robot Learning}, 2023, pp. 2105--2114.

\bibitem{yoshida2008game}
W.~Yoshida, R.~J. Dolan, and K.~J. Friston, ``Game theory of mind,'' \emph{PLoS Computational Biology}, p. e1000254, 2008.

\bibitem{doshi}
P.~Doshi, X.~Qu, A.~S. Goodie, and D.~L. Young, ``Modeling human recursive reasoning using empirically informed interactive partially observable markov decision processes,'' \emph{Transactions on Systems, Man, and Cybernetics - Part A: Systems and Humans}, pp. 1529--1542, 2012.

\bibitem{baker2017rational}
C.~L. Baker, J.~Jara-Ettinger, R.~Saxe, and J.~B. Tenenbaum, ``Rational quantitative attribution of beliefs, desires and percepts in human mentalizing,'' \emph{Nature Human Behaviour}, p. 0064, 2017.

\bibitem{bartlet_tom}
M.~E. Bartlett, C.~E. Edmunds, T.~Belpaeme, S.~Thill, and S.~Lemaignan, ``What can you see? identifying cues on internal states from the movements of natural social interactions,'' \emph{Frontiers in Robotics and AI}, p.~49, 2019.

\bibitem{mavrogiannis2019effects}
C.~Mavrogiannis, A.~M. Hutchinson, J.~Macdonald, P.~Alves-Oliveira, and R.~A. Knepper, ``Effects of distinct robot navigation strategies on human behavior in a crowded environment,'' in \emph{International Conference on Human-Robot Interaction}, 2019, pp. 421--430.

\bibitem{baker2014modeling}
C.~L. Baker and J.~B. Tenenbaum, ``Modeling human plan recognition using bayesian theory of mind,'' \emph{Plan, Activity, and Intent Recognition: Theory and Practice}, p.~86, 2014.

\bibitem{brooks2019building}
C.~Brooks and D.~Szafir, ``Building second-order mental models for human-robot interaction,'' \emph{arXiv preprint arXiv:1909.06508}, 2019.

\bibitem{mrs2}
L.~Bramblett, S.~Gao, and N.~Bezzo, ``Epistemic prediction and planning with implicit coordination for multi-robot teams in communication restricted environments,'' in \emph{International Conference on Robotics and Automation}, 2023, pp. 5744--5750.

\bibitem{machineToM}
N.~Rabinowitz, F.~Perbet, F.~Song, C.~Zhang, S.~M.~A. Eslami, and M.~Botvinick, ``Machine theory of mind,'' in \emph{International Conference on Machine Learning}, 2018, pp. 4218--4227.

\bibitem{NN-TOM}
A.-T. Nguyen, C.~Hieida, and T.~Nagai, ``A model of generating and predicting intention toward human-robot cooperation,'' in \emph{International Symposium on Robot and Human Interactive Communication}, 2018, pp. 113--120.

\bibitem{bayesiantom}
S.~Stacy, S.~Gong, A.~Parab, M.~Zhao, K.~Jiang, and T.~Gao, ``A bayesian theory of mind approach to modeling cooperation and communication,'' \emph{Computational Statistics}, p. e1631, 2024.

\bibitem{kingma2013auto}
D.~P. Kingma, ``Auto-encoding variational bayes,'' \emph{arXiv preprint arXiv:1312.6114}, 2013.

\bibitem{vqvae}
A.~Van Den~Oord, O.~Vinyals, \emph{et~al.}, ``Neural discrete representation learning,'' \emph{Advances in Neural Information Processing Systems}, 2017.

\bibitem{fsq}
F.~Mentzer, D.~Minnen, E.~Agustsson, and M.~Tschannen, ``Finite scalar quantization: Vq-vae made simple,'' \emph{arXiv preprint arXiv:2309.15505}, 2023.

\bibitem{skill-guided}
K.~Pertsch, Y.~Lee, Y.~Wu, and J.~J. Lim, ``Demonstration-guided reinforcement learning with learned skills,'' in \emph{Conference on Robot Learning}, 2022, pp. 729--739.

\bibitem{skill-prior}
K.~Pertsch, Y.~Lee, and J.~Lim, ``Accelerating reinforcement learning with learned skill priors,'' in \emph{Conference on robot learning}, 2021, pp. 188--204.

\bibitem{NEURIPS2021_a03caec5}
G.~Papoudakis, F.~Christianos, and S.~Albrecht, ``Agent modelling under partial observability for deep reinforcement learning,'' in \emph{Advances in Neural Information Processing Systems}, 2021.

\bibitem{sqvae}
Y.~Takida, T.~Shibuya, W.~Liao, C.-H. Lai, J.~Ohmura, T.~Uesaka, N.~Murata, S.~Takahashi, T.~Kumakura, and Y.~Mitsufuji, ``Sq-vae: Variational bayes on discrete representation with self-annealed stochastic quantization,'' in \emph{International Conference on Machine Learning}, 2022, pp. 20\,987--21\,012.

\bibitem{sac}
T.~Haarnoja, A.~Zhou, P.~Abbeel, and S.~Levine, ``Soft actor-critic: Off-policy maximum entropy deep reinforcement learning with a stochastic actor,'' in \emph{International conference on machine learning}, 2018, pp. 1861--1870.

\bibitem{interactiondataset}
W.~Zhan, L.~Sun, D.~Wang, H.~Shi, A.~Clausse, M.~Naumann, J.~K\"ummerle, H.~K\"onigshof, C.~Stiller, A.~de~La~Fortelle, and M.~Tomizuka, ``{INTERACTION} {Dataset}: {An} {INTERnational}, {Adversarial} and {Cooperative} {moTION} {Dataset} in {Interactive} {Driving} {Scenarios} with {Semantic} {Maps},'' \emph{arXiv:1910.03088 [cs, eess]}, 2019.

\end{thebibliography}

\end{document}